\documentclass[10pt,conference,letterpaper]{IEEEtran}

\usepackage[T1]{fontenc}

\usepackage{cite}
\usepackage{amsmath}
\usepackage{amssymb}
\usepackage{algorithm}
\usepackage{algpseudocode}
\usepackage{graphicx}
\usepackage{textcomp}
\usepackage{xcolor}
\usepackage{booktabs}
\usepackage{url}
\usepackage{authblk}
\usepackage{pgfplots}
\pgfplotsset{compat=1.18}

\def\BibTeX{{\rm B\kern-.05em{\sc i\kern-.025em b}\kern-.08em
  T\kern-.1667em\lower.7ex\hbox{E}\kern-.125emX}}

\newcommand{\method}{HASA} 

\usepackage{amssymb}
\usepackage{fancyhdr}
\fancypagestyle{firstpage}
{
    \fancyhead[L]{\footnotesize © 2026 IEEE.  Personal use of this material is permitted.  Permission from IEEE must be obtained for all other uses, in any current or future media, including reprinting/republishing this material for advertising or promotional purposes, creating new collective works, for resale or redistribution to servers or lists, or reuse of any copyrighted component of this work in other works. This paper is accepted at the 2026 IEEE International Conference on Edge Computing and Communications (IEEE EDGE 2026).}
    \fancyhead[R]{}
}
\begin{document}
\raggedbottom 

\title{HASA: Subnet Allocation for Compute-Constrained Model-Heterogeneous Federated Learning}


\author[1]{Amir Hossein Shahdadian}
\author[2]{Ahmed M. Abdelmoniem}
\author[3,4]{Mahdi Taheri} 
\author[5]{Samira Nazari}
\author[3]{Christian Herglotz}


\affil[1]{University of Naples "Federico II", Italy}
\affil[2]{Queen Mary University of London, United Kingdom}
\affil[3]{Brandenburg University of Technology Cottbus-Senftenberg, Germany}
\affil[4]{Tallinn University of Technology, Tallinn, Estonia}
\affil[5]{University of Zanjan, Iran}

\maketitle
\thispagestyle{firstpage}

\begin{abstract}
Edge services increasingly use federated learning to personalize on-device models while keeping sensitive data local. In practice, deployments must handle heterogeneity in both client resources and local data distributions. Model-heterogeneous federated learning lowers client cost by allowing each client to train a subnet of a shared supernet, but most subnet-allocation policies are driven by device constraints and do not explicitly account for statistical heterogeneity. This paper proposes Heterogeneity-Aware Subnet Allocation (HASA), a train-only rule that assigns subnet widths based on client heterogeneity scores computed from local training data while enforcing a fixed size-weighted compute budget. This design enables budget-matched comparisons with alternative allocation policies. On an article-title next-word prediction benchmark with seven clients, HASA improves unweighted mean client test accuracy over uniform allocation across 10 matched seeds, increasing mean client test accuracy from 13.82 percent to 14.32 percent, and improves worst-client accuracy on average. In a matched-budget comparison with representative partial-training baselines, HASA achieves the strongest worst-client and tail-client accuracy on this benchmark. A directionality ablation shows that assigning smaller subnets to more heterogeneous clients degrades both mean and tail performance. A cross-domain image-classification study further shows that the effectiveness of heterogeneity-aware allocation depends on how well the heterogeneity score reflects clients' need for additional model width.
\end{abstract}


\section{Introduction}

Edge computing has become a central paradigm for modern services that must satisfy latency, bandwidth, and data-locality requirements by moving computation closer to end users and data sources \cite{5, 6, 10}. Many such services increasingly use on-device learning, for example, in personalization tasks such as next-word prediction on mobile keyboards, where interaction data are privacy-sensitive and inherently distributed across devices \cite{bonawitz2019federatedlearningscaledesign}. Federated learning (FL) enables this setting by training a shared model through coordinated local updates without collecting raw data centrally \cite{pmlr-v54-mcmahan17a,hete2023study}. In large-scale deployments, privacy-preserving aggregation protocols such as secure aggregation are often used to further limit information exposure \cite{10.1145/3133956.3133982,bonawitz2019federatedlearningscaledesign}.

Despite these advantages, practical FL systems must address two forms of heterogeneity \cite{hete2023study,abdelmoniem2023refl}. First, \emph{system heterogeneity} arises from variability in client resources, including compute, memory, energy, and network conditions, which constrains how much local training each device can perform \cite{abdelmoniem2021aqfl}. Second, \emph{statistical heterogeneity} arises from non-IID and imbalanced data distributions across clients, which can induce client drift and slow or destabilize training under standard Federated Averaging (FedAvg) \cite{pmlr-v54-mcmahan17a}. Existing approaches such as FedProx and SCAFFOLD mitigate these effects by modifying optimization dynamics, but they primarily focus on stabilizing training of a shared global model rather than determining subnet-width distributions across heterogeneous clients \cite{MLSYS2020_1f5fe839,pmlr-v119-karimireddy20a}.

A complementary line of work addresses system heterogeneity through \emph{model-heterogeneous} FL, where clients train subnets of different widths within a shared supernet. Methods such as Federated Dropout and HeteroFL reduce client workload by training partial models and aggregating them into a global model \cite{caldas2019expandingreachfederatedlearning,diao2021heteroflcomputationcommunicationefficient}. Slimmable network frameworks further enable multiple subnet widths within a unified parameterization \cite{yu2018slimmableneuralnetworks}. However, in most existing systems, subnet allocation is driven primarily by device constraints, while statistical heterogeneity is handled separately through optimization or reweighting techniques.

This separation leaves subnet-width assignment largely independent of the statistical structure of client data in edge deployments with fixed training resources. Clients whose local data deviate more strongly from the global training mixture can require additional model width to achieve comparable accuracy, especially when the objective emphasizes client-level quality, including lower-tail performance, rather than only aggregate performance over examples.

\begin{figure*}[!t]
  \centering
  \includegraphics[width=\textwidth]{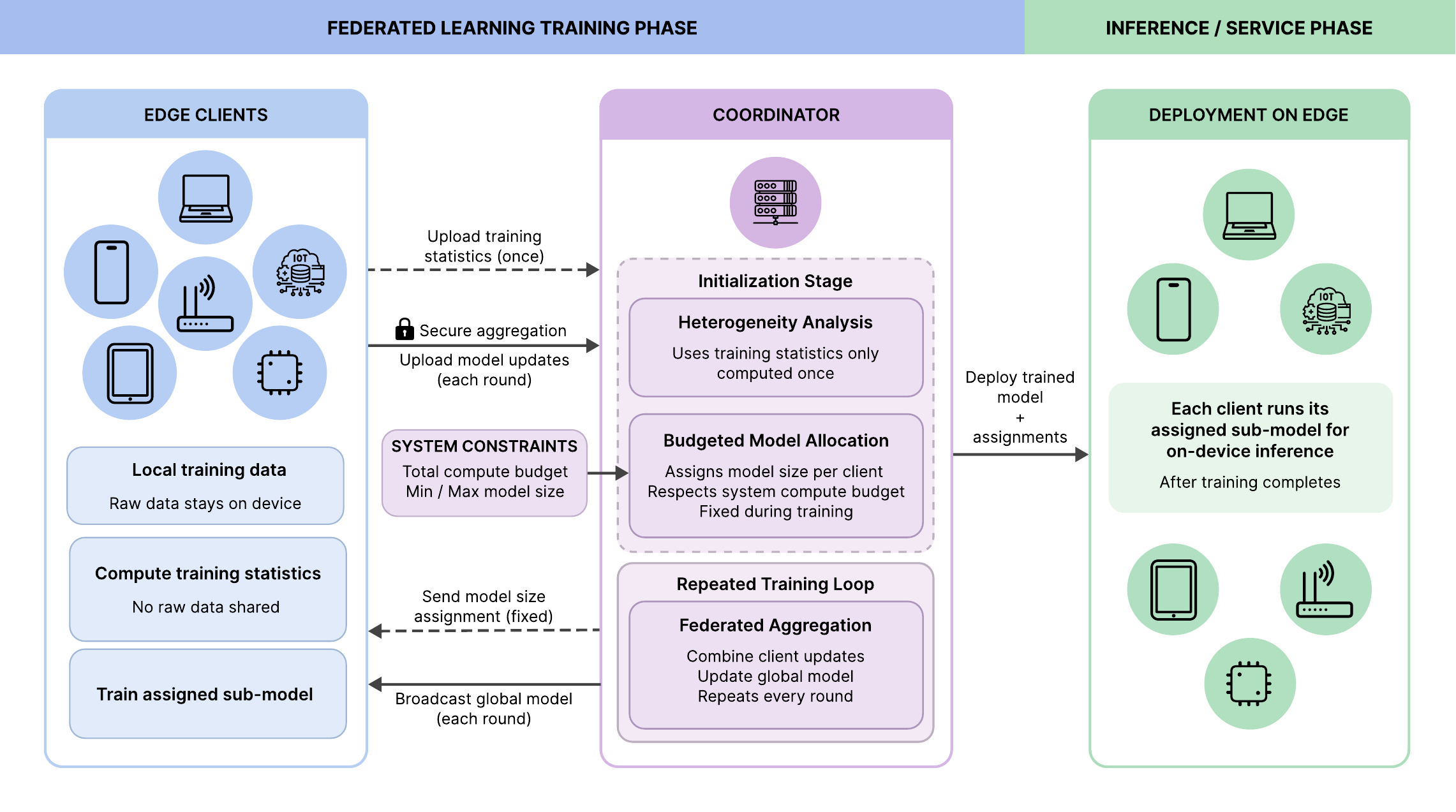}
  \caption{High-level overview of \method\ in an edge federated setting. Clients keep raw data local, compute train-only heterogeneity statistics, receive fixed subnet-width assignments under a shared budget, participate in federated training rounds, and use their assigned subnet for on-device inference.}
  \label{fig:arch}
\end{figure*}

To address this gap, \method\ (\emph{Heterogeneity-Aware Subnet Allocation}) assigns subnet widths in budgeted model-heterogeneous FL using train-only client heterogeneity scores. Each score is computed from the corresponding client's local training data. In the main instantiation, the score is the Jensen--Shannon divergence between local and global token distributions. The resulting allocations are normalized and rescaled to satisfy a fixed size-weighted compute budget, enabling budget-matched comparisons across allocation policies. This fixed-budget formulation isolates the effect of client-to-width assignment from increases in total training cost. Figure~\ref{fig:arch} summarizes the training and deployment workflow.

The evaluation covers three settings. On an article-title next-word prediction task, \method\ is compared with uniform allocation across 10 matched seeds under the same budget. A matched-budget benchmark then compares \method\ with representative model-heterogeneous partial-training baselines based on HeteroFL, Federated Dropout, and FedRolex, implemented on the same slimmable LSTM with a common training protocol. A cross-domain image-classification study evaluates how performance depends on the alignment between the heterogeneity score and clients' capacity demand.

The main contributions are:
\begin{itemize}
\item Formulation of model-heterogeneous federated learning under a fixed size-weighted compute budget, enabling fair budget-matched comparison across subnet-allocation policies.
\item Introduction of \method\ (\emph{Heterogeneity-Aware Subnet Allocation}), a train-only heterogeneity-aware allocation rule that maps client heterogeneity scores to subnet widths while enforcing a fixed global budget.
\item Development of a matched-budget evaluation framework integrating HeteroFL-style, FedDropout-style, and FedRolex-style partial-training strategies on a shared slimmable supernet under a unified training protocol.
\item Design of allocation-direction and cross-domain analyses to study the effect of heterogeneity-proxy alignment and subnet-width assignment under different non-IID conditions.
\end{itemize}

Section~II reviews related work. Section~III presents the problem formulation and evaluation metrics. Section~IV describes \method. Section~V details the experimental setup. Section~VI reports the results. Section~VII discusses limitations and practical considerations. Section~VIII concludes.

\section{Related Work}

Edge computing places computation closer to data sources to satisfy latency, bandwidth, and locality requirements in distributed services \cite{7488250,7807196}. Federated learning (FL) complements this setting by enabling collaborative model training while keeping raw data on devices or within organizations, which is particularly attractive in privacy- and regulation-sensitive applications \cite{li2025defense,bonawitz2019federatedlearningscaledesign}. In large-scale deployments, privacy-preserving aggregation protocols such as secure aggregation are often used so that the server can combine client updates without directly observing individual updates \cite{10.1145/3133956.3133982}. Practical edge FL systems, however, must simultaneously address two challenges: statistical heterogeneity, arising from non-IID client data~\cite{hete2023study}, and system heterogeneity, arising from differences in compute, memory, energy, and network resources~\cite{li2025prototypes, tam2025hierarchical}.

Statistical-heterogeneity methods primarily modify the optimization procedure. Non-IID data can cause client drift and unstable convergence under standard Federated Averaging (FedAvg) \cite{pmlr-v54-mcmahan17a}. FedProx stabilizes local training with a proximal regularization term designed for heterogeneous settings \cite{MLSYS2020_1f5fe839}, while SCAFFOLD reduces client drift using control variates \cite{pmlr-v119-karimireddy20a}. These methods improve optimization of a shared global model, but they do not directly address subnet-width distribution across clients under a fixed training cost.

System-heterogeneity methods reduce client workload by allowing clients to train smaller sub-models~\cite{li2025prototypes, tam2025hierarchical}. Federated Dropout reduces local computation by training only a subset of model parameters \cite{caldas2019expandingreachfederatedlearning}, while HeteroFL trains nested sub-networks of different widths and aggregates them into a shared global model \cite{diao2021heteroflcomputationcommunicationefficient}. FedRolex improves coverage of the server model over time through rolling sub-model extraction \cite{alam2023fedrolexmodelheterogeneousfederatedlearning}, and ScaleFL adapts model size to client resource constraints \cite{10204267}. Supernet-based approaches provide a practical basis for such methods: slimmable networks and once-for-all training support multiple widths within one shared parameterization \cite{yu2018slimmableneuralnetworks,cai2020onceforalltrainnetworkspecialize}, and SlimFL further considers communication and wireless aspects in this setting \cite{10004844}. These works mainly treat subnet width as a system-level decision tied to device capability.

Subnet extraction and subnet allocation are distinct design choices. A method can allow a client to train only part of the server model in each round while still requiring a policy for assigning larger and smaller subnets across clients. The allocation problem studied here uses train-only heterogeneity scores for subnet-width assignment under a fixed global budget. The matched-budget comparison against HeteroFL-style, FedDropout-style, and FedRolex-style baselines separates heterogeneity-aware width assignment from the implementation details of partial training.

Other lines of work address heterogeneity using mechanisms other than budgeted subnet-width assignment within a shared supernet. Distillation-based methods such as FedMD and FedDF enable collaboration among heterogeneous client models by aggregating predictions rather than averaging parameters \cite{li2019fedmdheterogenousfederatedlearning,lin2021ensembledistillationrobustmodel}. Fairness-oriented methods change the training objective to improve under-served clients; for example, agnostic federated learning (AFL) optimizes a worst-case client-mixture objective, and q-FFL modifies the federated objective to place more emphasis on clients with poorer local performance \cite{mohri2019agnosticfederatedlearning,li2020fairresourceallocationfederated}. Clustering-based methods such as IFCA learn multiple models for groups of clients with similar data distributions \cite{ghosh2021efficientframeworkclusteredfederated}. The allocation rule studied here maps a client heterogeneity score to subnet width under a fixed budget, allowing the client-to-width assignment to be analyzed without conflating it with increased total training cost.

\section{Problem Setup and Evaluation Metrics}

Consider $K$ clients. Client $i$ has training dataset $\mathcal{D}_i$ of size $n_i$ and local objective
\begin{equation}
F_i(w)=\mathbb{E}_{(x,y)\sim \mathcal{D}_i}\big[\ell(w;x,y)\big],
\label{eq:local_obj}
\end{equation}
where $x$ denotes an input example, $y$ its target output, $w$ the model parameters, and $\ell(w;x,y)$ the per-example loss. Let
\begin{equation}
N=\sum_{i=1}^{K} n_i .
\label{eq:total_size}
\end{equation}
The global objective minimized by Federated Averaging (FedAvg) is
\begin{equation}
F(w)=\sum_{i=1}^{K}\frac{n_i}{N}F_i(w).
\label{eq:global_obj}
\end{equation}

\subsection{Supernet and Client-Specific Subnets}
Let $w$ denote the parameters of a shared supernet. Client $i$ is assigned a subnet width ratio $r_i\in(0,1]$, which activates a fraction $r_i$ of the full model width and yields subnet $w_{r_i}$. In practice, allocations are restricted to bounds $[r_{\min},r_{\max}] \subset (0,1]$, specified later in the experimental setup. Each client trains only its assigned subnet, while the server aggregates client updates into the shared supernet.

\subsection{Fixed Compute Budget}
To compare subnet-allocation policies fairly, the formulation imposes a fixed \emph{budget level} $B$, defined as the client-size-weighted mean subnet width ratio:
\begin{equation}
B=\sum_{i=1}^{K}\frac{n_i}{N}r_i .
\label{eq:budget}
\end{equation}
For the article-title LSTM experiments (the main benchmark family), $B$ is fixed in advance, and all compared allocation policies must satisfy Eq.~\eqref{eq:budget}. Unless otherwise stated, full participation is assumed, so Eq.~\eqref{eq:budget} defines a per-round compute budget under size-proportional client weighting. The cross-domain CNN study uses a separate architecture-specific compute proxy, stated in Section~V-C, based on the size-weighted mean of $r_i^2$.

Communication savings require transmitting only the active subnet parameters. This is an implementation consequence of partial-model training, not part of the budget definition in Eq.~\eqref{eq:budget}.

\subsection{Evaluation Metrics}
Let $\mathrm{Acc}_i(r_i)$ denote the test accuracy of client $i$ when evaluated at its assigned subnet width $r_i$. The following client-level metrics are reported:
\begin{align}
\mathrm{MeanAcc} &= \frac{1}{K}\sum_{i=1}^{K}\mathrm{Acc}_i(r_i), \label{eq:meanacc}\\
\mathrm{WorstAcc} &= \min_{1\leq i\leq K}\mathrm{Acc}_i(r_i), \label{eq:worstacc}\\
\mathrm{P10Acc} &= Q_{0.1}\big(\{\mathrm{Acc}_i(r_i)\}_{i=1}^{K}\big), \label{eq:p10acc}
\end{align}
where $Q_{0.1}$ denotes the 10th percentile of the client-accuracy distribution.

For completeness, the size-weighted mean client accuracy is also defined as
\begin{equation}
\mathrm{WMeanAcc}=\sum_{i=1}^{K}\frac{n_i}{N}\mathrm{Acc}_i(r_i),
\label{eq:wmeanacc}
\end{equation}
but it is not used as a headline metric because the goal is to improve client-level service quality across clients rather than average performance over examples.

The deployment target is the per-client accuracy of the assigned subnet under the fixed budget in Eq.~\eqref{eq:budget}. Accordingly, each client is evaluated at its allocated width $r_i$, not at the full supernet width.

\section{HASA: Heterogeneity-Aware Subnet Allocation}

\begin{figure*}[t]
\centering
\includegraphics[width=\textwidth]{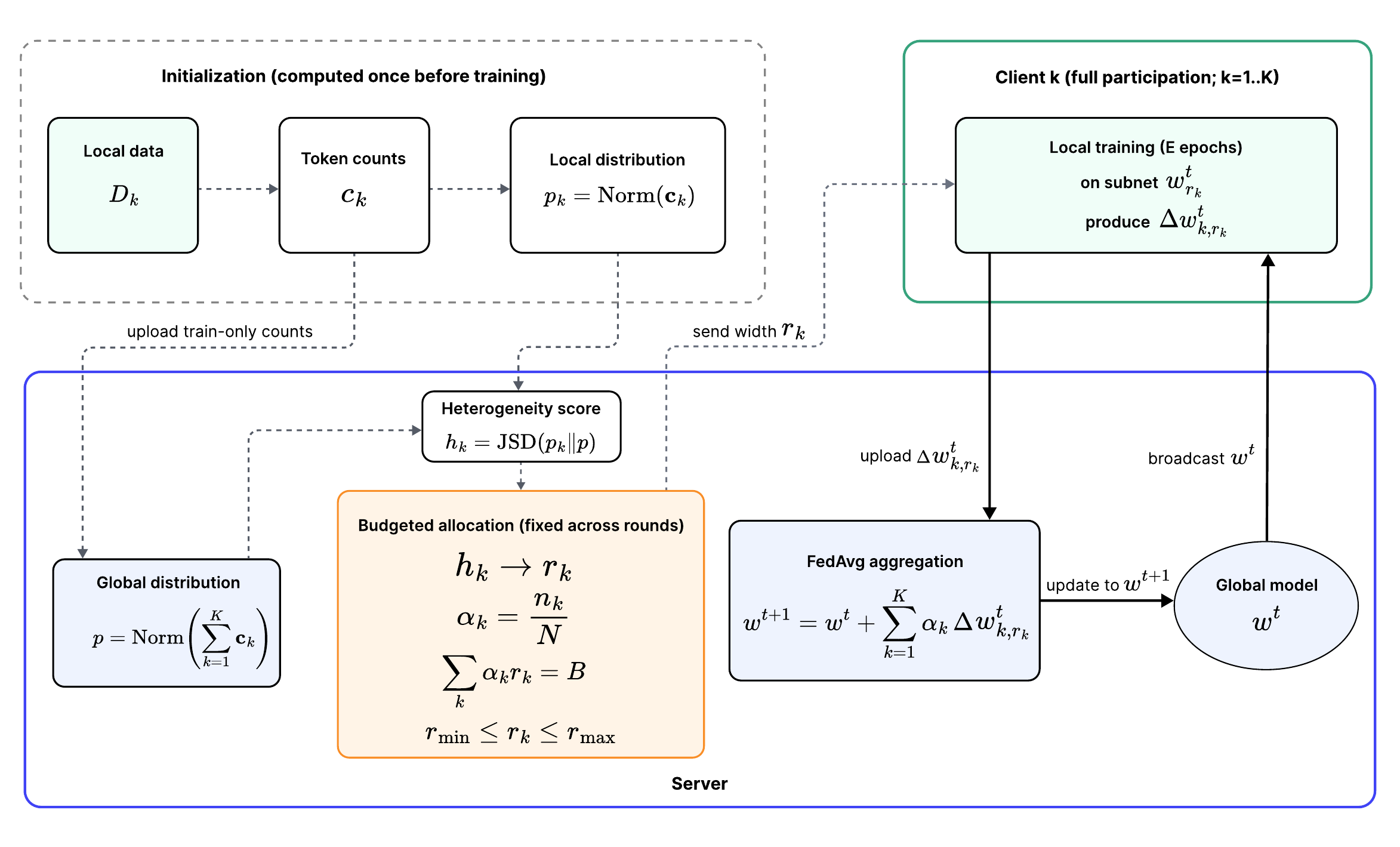}
\caption{Overview of \method: train-only heterogeneity scoring, budgeted width mapping, and federated training with client-specific subnets.}
\label{fig:hasa_overview}
\end{figure*}

\method\ comprises train-only heterogeneity scoring, score-to-width mapping, and fixed-budget projection. Figure~\ref{fig:hasa_overview} summarizes the workflow before the detailed formulation.

\subsection{Train-Only Heterogeneity Score}
HASA assigns each client a \emph{heterogeneity score} $h_i$ computed from training data only. In the main text application, client $i$ is represented by a smoothed unigram token distribution $p_i$ over a shared vocabulary $\mathcal{V}$, and this distribution is compared with the global training distribution $p_{\mathrm{g}}$.

For each token $v\in\mathcal{V}$,
\begin{align}
p_i(v) &= \frac{c_{i,v}+\alpha}{\sum_{u\in\mathcal{V}}(c_{i,u}+\alpha)}, \label{eq:pi}\\
p_{\mathrm{g}}(v) &= \frac{c_{v}+\alpha}{\sum_{u\in\mathcal{V}}(c_{u}+\alpha)}, \label{eq:pg}
\end{align}
where $c_{i,v}$ is the count of token $v$ in client $i$'s training data, $c_v$ is the corresponding pooled count over all clients, and $\alpha>0$ is a smoothing constant.

The heterogeneity score is then defined as the Jensen--Shannon divergence (JSD) between $p_i$ and $p_{\mathrm{g}}$:
\begin{equation}
h_i=\mathrm{JSD}(p_i\|p_{\mathrm{g}})
=\frac{1}{2}\mathrm{KL}(p_i\|m_i)+\frac{1}{2}\mathrm{KL}(p_{\mathrm{g}}\|m_i),
\label{eq:jsd}
\end{equation}
where
\begin{equation}
m_i=\frac{1}{2}\big(p_i+p_{\mathrm{g}}\big)
\label{eq:mixture}
\end{equation}
and $\mathrm{KL}(\cdot\|\cdot)$ denotes the Kullback--Leibler divergence. The logarithm base used inside $\mathrm{KL}$ only rescales the score and does not affect \method\ because the subsequent allocation step is rank-based.

Statistical heterogeneity can increase the mismatch between a client's local objective and the global training objective. Under FedAvg-style training, such a mismatch is one source of client drift and unstable convergence \cite{pmlr-v54-mcmahan17a,pmlr-v119-karimireddy20a}. A client whose token distribution is far from the global mixture can require additional representational capacity to model domain-specific lexical patterns while still sharing parameters through the global supernet. \method\ treats the heterogeneity score as a capacity-demand proxy: higher-scoring clients are assigned no less width on average, subject to the global budget and any client-specific feasibility caps. This allocation assumption is evaluated empirically rather than imposed as generally valid; the directionality ablation and the CIFAR-10 diagnostic in Section~VI test cases in which the proxy is aligned or misaligned. 

The one-time initialization phase in \method\ computes the global reference distribution. Clients do not upload raw text; they compute token-count vectors locally, and the server obtains the pooled counts $\{c_v\}_{v\in\mathcal{V}}$ by summation. In privacy-sensitive deployments, this summation can be implemented with secure aggregation so that the server observes only the aggregate count vector, not individual histograms \cite{10.1145/3133956.3133982}. In the article-title benchmark, the 3{,}437-token model vocabulary implies about 13.4~KB per dense 32-bit client count vector before secure-aggregation overhead, paid once before training. For larger vocabularies, sparse histograms, hashing, top-$k$ counts, or differentially private noisy counts can reduce disclosure or bandwidth at the cost of a noisier proxy.

\subsection{Score Normalization, Width Mapping, and Budget Enforcement}
Let $\mathrm{rank}(h_i)\in\{1,\dots,K\}$ denote the rank of client $i$'s heterogeneity score among $\{h_j\}_{j=1}^{K}$, where rank 1 is the smallest score and ties receive average rank. These ranks are converted to normalized scores in $[0,1]$:
\begin{equation}
\tilde{h}_i=\frac{\mathrm{rank}(h_i)-1}{K-1}.
\label{eq:ranknorm}
\end{equation}

The normalized scores are then mapped to preliminary subnet widths:
\begin{equation}
\hat{r}_i^{(0)} = r_{\min} + (r_{\max}-r_{\min})\tilde{h}_i .
\label{eq:init_width}
\end{equation}

To enforce the fixed budget in Eq.~\eqref{eq:budget}, these preliminary widths are iteratively scaled and projected. Let
\begin{equation}
\mathcal{P}_{[a,b]}(x)=\min(\max(x,a),b)
\label{eq:projection}
\end{equation}
denote projection onto the interval $[a,b]$. At iteration $t$, define the scaling factor
\begin{equation}
s^{(t)}=\frac{B}{\sum_{j=1}^{K}(n_j/N)\hat{r}_j^{(t)}} .
\label{eq:scale}
\end{equation}
The updated widths are
\begin{equation}
\hat{r}_i^{(t+1)}=
\mathcal{P}_{[r_{\min},u_i]}\!\left(s^{(t)}\hat{r}_i^{(t)}\right),
\qquad i=1,\dots,K .
\label{eq:budget_project}
\end{equation}
After $T$ iterations, the final allocation is
\begin{equation}
r_i=\hat{r}_i^{(T)}.
\label{eq:final_alloc}
\end{equation}
In the implementation, two scale-and-project passes are used. Because widths are clamped and later discretized into integer active-unit counts, small residual deviations from the nominal budget remain; realized budget values are therefore reported explicitly in the result tables.

Algorithm~\ref{alg:hasa} summarizes the procedure.

\begin{algorithm}[t]
\caption{HASA allocation under a fixed size-weighted budget and client caps}
\label{alg:hasa}
\begin{algorithmic}[1]
\setlength{\itemsep}{3pt}   
\Statex \textbf{Input:} client sizes $\{n_i\}_{i=1}^{K}$, heterogeneity scores $\{h_i\}_{i=1}^{K}$, bounds $[r_{\min},r_{\max}]$, client caps $\{u_i\}_{i=1}^{K}$, budget $B$, number of iterations $T$
\Statex \textbf{Output:} allocated subnet width ratios $\{r_i\}_{i=1}^{K}$
\State Compute normalized scores $\tilde{h}_i \leftarrow (\mathrm{rank}(h_i)-1)/(K-1)$ for all $i$
\State Initialize $\hat{r}_i^{(0)} \leftarrow r_{\min} + (r_{\max}-r_{\min})\tilde{h}_i$ for all $i$
\For{$t=0$ to $T-1$}
    \State $s^{(t)} \leftarrow \dfrac{B}{\sum_{j=1}^{K}(n_j/N)\hat{r}_j^{(t)}}$
    \For{$i=1$ to $K$}
        \State $\hat{r}_i^{(t+1)} \leftarrow \mathcal{P}_{[r_{\min},u_i]}\!\left(s^{(t)}\hat{r}_i^{(t)}\right)$
    \EndFor
\EndFor
\State \Return $r_i \leftarrow \hat{r}_i^{(T)}$ for all $i$
\end{algorithmic}
\end{algorithm}

\subsection{Baselines and Ablations}
All baselines use the same bounds, client caps, and budget-enforcement procedure so that only the allocation rule changes.

Under the uniform allocation, all clients receive the same subnet width. In settings where $B \in [r_{\min},r_{\max}]$, this gives
\begin{equation}
r_i = B \qquad \text{for all } i.
\label{eq:uniform}
\end{equation}

The size-only allocation replaces the heterogeneity score with a min--max normalized size score. Let
\begin{equation}
\tilde{n}_i=
\begin{cases}
\dfrac{n_i-\min_j n_j}{\max_j n_j-\min_j n_j}, & \max_j n_j>\min_j n_j,\\[4pt]
0.5, & \max_j n_j=\min_j n_j,
\end{cases}
\label{eq:size_score}
\end{equation}
The normalized size score $\tilde{n}_i$ is mapped to subnet widths using the same procedure as Eqs.~\eqref{eq:init_width}--\eqref{eq:final_alloc}.

The mixed allocation combines the min--max normalized size score and the rank-normalized heterogeneity score:
\begin{equation}
z_i=\gamma \tilde{n}_i + (1-\gamma)\tilde{h}_i,\qquad \gamma\in[0,1],
\label{eq:mixed_score}
\end{equation}
The combined score $z_i$ is mapped to subnet widths using the same width-mapping and budget-enforcement procedure.

The inverse ablation reverses the heterogeneity-to-width mapping by replacing $\tilde{h}_i$ with
\begin{equation}
\tilde{h}_i^{\mathrm{inv}} = 1-\tilde{h}_i,
\label{eq:inverse_score}
\end{equation}
The same width-mapping and budget-enforcement procedure is then applied. This ablation measures sensitivity to allocation direction under the same fixed budget.

\section{Experimental Setup}

Two evaluation protocols are used on the article-title benchmark. The 10-seed main comparison against a uniform allocation uses the original FedAvg pipeline, in which the server aggregates the full parameter tensor after each round. The matched-budget comparison against external partial-training baselines instead uses \emph{selective aggregation}, because those baselines update only the parameter entries touched by the active subnet in a given round. This distinction is kept explicit throughout the paper: the first protocol measures the effect of heterogeneity-aware subnet allocation in the original training pipeline, whereas the second isolates the allocation rule against representative partial-training baselines under a fair aggregation rule.

\subsection{Article-Title Next-Word Prediction}

The next-word prediction benchmark is constructed from Medium article titles, with clients partitioned by publication ($K{=}7$). Titles are obtained from the \emph{Medium Articles Dataset} hosted on Kaggle~\cite{medium2019kaggle}, which provides article metadata including title and publication fields. A total of 5\% of articles is reserved as a global out-of-distribution (OOD) test set (319 articles; 2{,}285 sequences). From the remaining titles, seven publications are selected as clients to span a wide range of dataset sizes, including one very small client for lower-tail evaluation; a global in-federation test set sampled from these publications contains 2{,}143 sequences. Each client is split at the \emph{article level} into 70/10/20 train/validation/test to avoid leakage across splits. Titles are lowercased and tokenized, the vocabulary is built from the non-OOD pool, and heterogeneity scores are computed from training splits only. Sliding variable-length $n$-grams over titles (with $n \leq 23$, the maximum title length) yields 28{,}248 training sequences across clients. Client totals in articles are: The Startup (2{,}686), Towards Data Science (1{,}298), Data Driven Investor (689), UX Collective (503), The Writing Cooperative (357), Better Marketing (216), and Better Humans (25).

The model is a slimmable recurrent language model with embedding dimension 128 and hidden dimension 256, implemented as a one-layer Long Short-Term Memory (LSTM) network. A subnet width ratio $r$ activates the first $\lfloor r \cdot 256 \rfloor$ hidden units and the corresponding sliced weights, yielding a client-specific subnet embedded in a shared supernet.

Unless otherwise stated, the article-title experiments use full participation, meaning that all $K$ clients participate in every federated round. Training runs for 50 FL rounds with 1 local epoch per round using Adam (batch size 64; learning rate $10^{-3}$). Dropout is disabled and weight decay is set to zero. Each client trains only its assigned subnet throughout training; multiple widths are not sampled for the same client within a round. Test metrics are reported at the final round and are additionally logged every 5 rounds during training. For all article-title experiments, the width bounds are set to $r_{\min}=0.2$ and $r_{\max}=0.8$ and a fixed size-weighted budget level $B=0.5$. These bounds avoid unstable very small subnets and near-full-width subnets while allowing meaningful reallocation around the uniform baseline.

This benchmark represents a controlled edge-node or cross-silo study rather than a large-scale cross-device deployment. Full participation isolates subnet allocation from client sampling, availability bias, stragglers, and device-specific feasibility effects \cite{hete2023study,bonawitz2019federatedlearningscaledesign}.

\subsection{Matched-Budget Baseline Comparison}

To compare \method\ with representative baselines on the same article-title benchmark, four matched-budget methods are evaluated using the same slimmable LSTM. The methods are Uniform-Static, HeteroFL-style, FedDropout-style, and FedRolex-style. Uniform-Static trains each client with one shared subnet width and static prefix extraction. The HeteroFL-style baseline uses a size-based width rule with static prefix extraction. The FedDropout-style baseline keeps the same size-based rule but samples a random set of active hidden units each round. The FedRolex-style baseline also keeps the size-based rule and rotates a contiguous hidden-unit window across rounds. \method\ uses the heterogeneity-aware width assignment from Section~IV with static extraction.

All methods in this comparison use the same client split, optimizer, number of rounds, local epochs, width bounds, and size-proportional client aggregation weights. Because this benchmark does not include measured device-capability traces, all external baselines use the same size-driven subnet-allocation rule. This keeps the comparison focused on heterogeneity-aware versus size-driven width assignment rather than on device scheduling.

For this baseline comparison, \emph{selective aggregation} is used: only parameter entries updated by a client's active subnet are averaged at the server, while untouched entries retain their previous global value. This avoids diluting parameters that were not trained in that round and is therefore the appropriate aggregation rule for partial-training baselines. The nominal budget target remains $B=0.5$, but realized budget values can differ slightly after converting width ratios to integer hidden-unit counts and projecting them back into the feasible range. Realized overhead values are therefore reported in addition to nominal budget settings.

Along with client-level mean, worst-client, and P10 accuracy, the reported modeled-overhead metrics include realized weighted forward-MAC ratio and simulated active-submodel uplink per round. Simulated communication is computed from active parameter count multiplied by four bytes per parameter and is interpreted as a controlled communication proxy rather than measured network traffic. \method's allocation is fixed after initialization; it does not continuously adapt subnet widths during training, and the width assignment itself is negligible compared with model-update traffic.

\subsection{Image-Classification Cross-Domain Study}

A CIFAR-10 benchmark experiment is included as a cross-domain diagnostic of whether a train-only heterogeneity score aligns with client capacity demand outside the text domain. CIFAR-10 comprises 10 object classes of small color images, making it suitable for assessing allocation behavior in a different data modality.

CIFAR-10 is partitioned into 5 clients using constrained Dirichlet-based non-IID sampling with concentration parameter $\alpha_{\mathrm{Dir}}=0.5$: per-client class proportions are sampled from a Dirichlet distribution, and individual images are assigned to clients according to those sampled shares, while enforcing at least 500 images and at least 3 distinct classes per client. Each client is then split 80/20 into local train/test sets at the image level.

In the CIFAR-10 experiment, the train-only heterogeneity score is the mean per-image pixel standard deviation averaged over each client's training set, referred to as a \emph{visual-complexity score}. This unlabeled proxy can be computed without class labels but does not directly capture label-distribution skew: two clients can have similar pixel variance while concentrating on different classes. Label-distribution JSD provides a more direct proxy when labels are available, while frozen-encoder feature distances provide an unlabeled alternative. The experiment is a proxy-alignment diagnostic rather than evidence that pixel variance is a generally suitable heterogeneity score. A slimmable CNN is used with three convolutional blocks (base channels 64) and two fully connected layers (hidden size 256), where width ratio $r$ directly sets the active number of channels in each convolutional block. Training runs for 50 rounds with 3 local epochs per round, batch size 64, base learning rate 0.005, and multiplicative learning-rate decay 0.99. Width ratios are restricted to $r \in [0.4,1.0]$.

Because convolutional compute scales approximately with the square of width, this CNN study uses an architecture-specific budget proxy: the size-weighted mean of $r_i^2$ is kept fixed at 0.25, which corresponds to the uniform baseline at $r=0.5$. This replaces Eq.~\eqref{eq:budget} for the CNN experiment only. Selective aggregation is again used over updated parameter entries: for parameter element $j$, let $m_{ij}\in\{0,1\}$ indicate whether client $i$ updated that element under its assigned subnet. The server then aggregates as
\begin{equation}
w^{t+1}_j =
\frac{\sum_i (n_i/N)\, m_{ij}\, w^{t+1}_{i,j}}
{\sum_i (n_i/N)\, m_{ij}}.
\end{equation}

In addition to average and worst-client accuracy evaluated at each client's allocated subnet width, global test accuracy at the common width ratio $r=0.5$ is also reported. This common-width evaluation is included only as a diagnostic so that policies with different allocations can also be compared under the same inference-time width.

\section{Results}

\subsection{Main Results on the Article-Title Benchmark}

A 10-seed matched comparison between Uniform and \method\ under the same fixed budget ($B{=}0.5$) in the original FedAvg pipeline provides the main article-title result. Matched seeds use the same 10 random seeds for both policies, enabling pairwise comparison of run-level differences.

\begin{figure}[t]
\centering
\begin{tikzpicture}
\begin{axis}[
    ybar,
    bar width=8pt,
    width=\columnwidth,
    height=0.55\columnwidth,
    ymin=10, ymax=16,
    ylabel={Accuracy (\%)},
    symbolic x coords={Mean,Worst,P10},
    xtick=data,
    legend style={at={(0.5,1.05)},anchor=south,legend columns=2,font=\small},
    error bars/y dir=both,
    error bars/y explicit,
    nodes near coords style={font=\tiny},
]
\addplot+ coordinates {(Mean,13.82) +- (0,0.40) (Worst,11.47) +- (0,0.56) (P10,12.04) +- (0,0.36)};
\addplot+ coordinates {(Mean,14.32) +- (0,0.36) (Worst,11.99) +- (0,0.70) (P10,12.52) +- (0,0.37)};
\legend{Uniform,\method}
\end{axis}
\end{tikzpicture}
\caption{Client-level test accuracy on the article-title benchmark over 10 matched seeds: mean with $\pm 1$ standard deviation error bars. All metrics are computed on each client's assigned subnet width under fixed budget $B{=}0.5$.}
\label{fig:main_bar}
\end{figure}
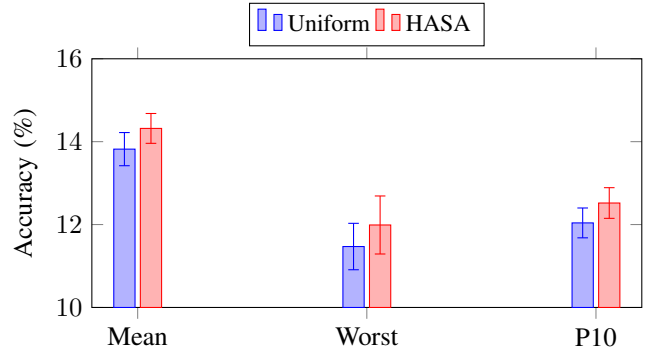

Figure~\ref{fig:main_bar} summarizes the results. Across 10 matched seeds, \method\ improves mean client test accuracy from 13.82\% to 14.32\%. The average seed-wise paired improvement is 0.49 percentage points, while the rounded endpoint means differ by 0.50 points. Under the directional hypothesis of improvement over Uniform, one-sided paired tests on the per-seed mean accuracies reject the null of no improvement (paired $t$-test: $t{=}4.74$, $p{<}0.001$; Wilcoxon signed-rank test: $p{=}0.003$; Cohen's $d{=}1.23$).

Tail metrics also improve on average: worst-client accuracy increases from 11.47\% to 11.99\%, and P10 increases from 12.04\% to 12.52\%. The P10 improvement is statistically significant ($t{=}2.57$, $p{=}0.015$, $d{=}1.36$). The worst-client improvement does not reach the $\alpha{=}0.05$ level under a one-sided paired $t$-test ($t{=}1.56$, $p{=}0.077$, $d{=}0.83$), reflecting higher variance across seeds. The mean-accuracy and P10 gains are therefore treated as the main statistically supported results, with the worst-client improvement interpreted as directionally consistent but less certain evidence. Mean test perplexity also decreases from 521.1 ($\pm$28.5) to 513.3 ($\pm$25.5) under \method, consistent with improved language-modeling quality. These gains are modest in absolute terms, but they are obtained by reallocating a fixed subnet budget rather than increasing total size-weighted width.

Table~\ref{tab:main_10seed} reports the full 10-seed results for all allocation policies evaluated in the original FedAvg pipeline. Among the budgeted methods, \method\ achieves the best mean, worst-client, and P10 accuracy. The Size-only policy, which allocates larger subnets to clients with more data, is the weakest budgeted method on all client-level metrics. The Mixed policy ($\gamma{=}0.5$), which combines size and heterogeneity scores, falls between Uniform and \method. FedAvg-Full, which trains the full supernet on all clients without a budget constraint, is included as a reference point; it does not outperform \method\ on any client-level metric in this pipeline, though it was not separately tuned for full-width training. The paired statistical tests reported above use a dedicated 10-seed pairwise run; Table~\ref{tab:main_10seed} reports a separate 10-seed evaluation of all five policies under the same protocol.

\begin{table}[t]
\centering
\caption{Ten-seed comparison of allocation policies on the article-title benchmark under the original FedAvg pipeline ($B{=}0.5$). All client-level metrics are computed at each client's assigned subnet width. Best budgeted result in \textbf{bold}.}
\label{tab:main_10seed}
\scriptsize
\setlength{\tabcolsep}{3.0pt}
\renewcommand{\arraystretch}{1.3}
\begin{tabular*}{\columnwidth}{@{\extracolsep{\fill}}lccccc@{}}
\toprule
Policy & Mean & Worst & P10 & Avg PPL & Wtd Ratio \\
 & (\%) & (\%) & (\%) &  & (\%) \\
\midrule
\method\ & \textbf{14.32}{\scriptsize$\pm$0.36} & \textbf{11.99}{\scriptsize$\pm$0.70} & \textbf{12.52}{\scriptsize$\pm$0.37} & \textbf{513.3}{\scriptsize$\pm$25.5} & 49.6 \\
Uniform & 13.82{\scriptsize$\pm$0.40} & 11.47{\scriptsize$\pm$0.56} & 12.04{\scriptsize$\pm$0.36} & 520.8{\scriptsize$\pm$28.3} & 50.0 \\
Mixed ($\gamma{=}0.5$) & 13.60{\scriptsize$\pm$0.60} & 10.56{\scriptsize$\pm$0.64} & 11.32{\scriptsize$\pm$0.60} & 559.5{\scriptsize$\pm$25.2} & 50.0 \\
Size-only & 12.74{\scriptsize$\pm$0.48} & 10.70{\scriptsize$\pm$0.80} & 10.95{\scriptsize$\pm$0.66} & 540.7{\scriptsize$\pm$13.1} & 50.0 \\
\midrule
FedAvg-Full (ref.) & 13.03{\scriptsize$\pm$0.50} & 10.96{\scriptsize$\pm$0.81} & 11.51{\scriptsize$\pm$0.48} & 576.8{\scriptsize$\pm$35.0} & 100.0 \\
\bottomrule
\end{tabular*}
\end{table}

\subsection{Comparison with Model-Heterogeneous FL Baselines}

Representative model-heterogeneous partial-training baselines are evaluated under the same nominal budget target $B{=}0.5$. Because these baselines update only the parameter entries touched by the active subnet, this comparison uses \emph{selective aggregation} rather than the original full-tensor FedAvg pipeline. All methods share the same client split, optimizer, number of rounds, local epochs, width bounds, and size-proportional client weighting.

Table~\ref{tab:hetero_baselines} reports the 10-seed results. Among the budgeted methods, \method\ achieves the strongest worst-client accuracy (11.73\%) and P10 (12.22\%). On mean accuracy, \method\ (13.81\%) and Uniform-Static (13.94\%) are statistically indistinguishable ($\Delta{=}{-}0.13\%$, $t{=}{-}0.68$, $p{=}0.74$), while both significantly outperform the size-driven baselines. Paired one-sided $t$-tests confirm that \method\ improves mean accuracy over HeteroFL-style ($\Delta{=}{+}1.22\%$, $t{=}5.62$, $p{<}0.001$), FedDropout-style ($\Delta{=}{+}1.58\%$, $t{=}9.66$, $p{<}0.001$), and FedRolex-style ($\Delta{=}{+}1.35\%$, $t{=}6.98$, $p{<}0.001$). These differences are not explained by materially larger modeled cost: realized weighted forward-MAC ratios lie between 34.81\% and 36.98\%, and simulated per-round uplink differs by at most 0.03~MB. HeteroFL-style, FedDropout-style, and FedRolex-style have identical overheads because they use the same size-driven widths and differ only in extraction pattern.

\begin{table}[t]
\centering
\caption{Ten-seed matched-budget comparison on the article-title benchmark under selective aggregation ($B{=}0.5$). Communication is simulated active-submodel traffic. Best budgeted value in each client-level metric column is shown in \textbf{bold}.}
\label{tab:hetero_baselines}
\scriptsize
\setlength{\tabcolsep}{2.8pt}
\renewcommand{\arraystretch}{1.2}
\begin{tabular*}{\columnwidth}{@{\extracolsep{\fill}}lccccc@{}}
\toprule
Method & Mean & Worst & P10 & Wtd MACs & Sim UL \\
 & (\%) & (\%) & (\%) & (\%) & (MB) \\
\midrule
\method\ & 13.81{\scriptsize$\pm$0.39} & \textbf{11.73}{\scriptsize$\pm$0.73} & \textbf{12.22}{\scriptsize$\pm$0.62} & 36.98 & 4.08 \\
Uniform-Static & \textbf{13.94}{\scriptsize$\pm$0.37} & 11.49{\scriptsize$\pm$0.75} & 11.97{\scriptsize$\pm$0.66} & 34.81 & 4.06 \\
HeteroFL-style & 12.59{\scriptsize$\pm$0.52} & 10.02{\scriptsize$\pm$1.13} & 10.52{\scriptsize$\pm$0.60} & 36.85 & 4.09 \\
FedRolex-style & 12.46{\scriptsize$\pm$0.36} & 10.25{\scriptsize$\pm$0.44} & 10.60{\scriptsize$\pm$0.32} & 36.85 & 4.09 \\
FedDropout-style & 12.23{\scriptsize$\pm$0.55} & 9.03{\scriptsize$\pm$1.07} & 9.61{\scriptsize$\pm$0.74} & 36.85 & 4.09 \\
\midrule
FedAvg-Full (ref.) & 13.01{\scriptsize$\pm$0.56} & 10.70{\scriptsize$\pm$0.85} & 11.25{\scriptsize$\pm$0.70} & 100.00 & 6.87 \\
\bottomrule
\end{tabular*}
\end{table}

The advantage of \method\ over Uniform-Static appears primarily in worst-client and P10 accuracy rather than in mean accuracy. Under selective aggregation, Uniform-Static is competitive on mean accuracy, but \method\ improves worst-client accuracy by 0.24 percentage points and P10 by 0.25 percentage points on average. Among the size-driven model-heterogeneous baselines, FedDropout-style exhibits the weakest client-level metrics---particularly on the worst client (9.03\%)---consistent with random subnet extraction disrupting learned feature co-adaptation across rounds.

\subsection{Per-Client Diagnostic Analysis}

\begin{table}[t]
\centering
\caption{Per-client diagnostic over 10 matched seeds (original FedAvg pipeline, $B{=}0.5$). Size is the number of training sequences. JSD is the Jensen--Shannon divergence of each client's token distribution from the global distribution. Width denotes the subnet width ratio assigned by \method. $\Delta$ is the mean accuracy change from Uniform to \method.}
\label{tab:per_client}
\begingroup
\scriptsize
\setlength{\tabcolsep}{2.5pt}
\renewcommand{\arraystretch}{1.1}
\begin{tabular*}{\columnwidth}{@{\extracolsep{\fill}}lrrcrrc@{}}
\toprule
Client & Size & JSD & Width & Uniform & \method & $\Delta$ \\
 & & & (\%) & (\%) & (\%) & (pp) \\
\midrule
Towards Data Sci.\ & 6054 & 0.110 & 73.1 & 14.62{\scriptsize$\pm$0.71} & 16.80{\scriptsize$\pm$0.45} & +2.18 \\
UX Collective & 2570 & 0.120 & 80.0 & 11.94{\scriptsize$\pm$0.89} & 13.93{\scriptsize$\pm$0.89} & +1.99 \\
Data Driven Inv.\ & 3354 & 0.096 & 43.8 & 11.99{\scriptsize$\pm$0.58} & 12.46{\scriptsize$\pm$0.59} & +0.47 \\
The Startup & 13215 & 0.043 & 29.2 & 13.26{\scriptsize$\pm$0.47} & 13.71{\scriptsize$\pm$0.43} & +0.45 \\
Better Marketing & 1195 & 0.105 & 58.4 & 14.36{\scriptsize$\pm$0.73} & 14.10{\scriptsize$\pm$1.09} & $-$0.26 \\
Writing Coop.\ & 1719 & 0.121 & 80.0 & 16.17{\scriptsize$\pm$0.77} & 15.41{\scriptsize$\pm$0.51} & $-$0.76 \\
Better Humans & 141 & 0.192 & 80.0 & 14.73{\scriptsize$\pm$1.90} & 13.82{\scriptsize$\pm$2.18} & $-$0.91 \\
\bottomrule
\end{tabular*}
\endgroup
\end{table}

Table~\ref{tab:per_client} disaggregates the 10-seed main comparison into per-client outcomes. The largest accuracy gains under \method\ occur at Towards Data Science ($+2.18$ pp) and UX Collective ($+1.99$ pp), both of which receive substantially wider subnets than under the uniform allocation (73.1\% and 80.0\% versus the uniform 50.0\%). The Startup, the largest client, receives the narrowest subnet (29.2\%) yet still gains $+0.45$ pp, suggesting that its low heterogeneity (JSD$=$0.043) is adequately served by a smaller subnet.

Three clients show small negative deltas: Better Marketing ($-0.26$ pp), The Writing Cooperative ($-0.76$ pp), and Better Humans ($-0.91$ pp). The Writing Cooperative and Better Humans both receive the maximum width (80.0\%) under \method, indicating that their slight regressions are not caused by width reduction. Better Humans is the smallest client (141 training sequences), and its high variance ($\pm$2.18\%) reflects the inherent instability of evaluation on a very small test set. This outcome illustrates a limitation of pure rank-based scoring: very small clients can receive extreme heterogeneity ranks because their empirical token distributions are noisy. Smoothing and width caps partly mitigate this issue; small-sample score shrinkage toward the population mean before ranking is a possible extension.

The net effect of these per-client tradeoffs is positive: the aggregate mean-accuracy gain of $+0.45$ pp across clients (computed from Table~\ref{tab:per_client}) is driven by large improvements at mid-to-high heterogeneity clients that outweigh small losses elsewhere.

\subsection{Allocation-Direction Ablation}

\begin{table}[t]
\centering
\caption{Allocation-direction ablation on the article-title benchmark over 10 matched seeds (original FedAvg pipeline, $B{=}0.5$). Best budgeted result in \textbf{bold}.}
\label{tab:directionality}
\scriptsize
\setlength{\tabcolsep}{3.5pt}
\renewcommand{\arraystretch}{1.5}
\begin{tabular*}{\columnwidth}{@{\extracolsep{\fill}}lcccc@{}}
\toprule
Policy & Mean (\%) & Worst (\%) & P10 (\%) & Avg PPL \\
\midrule
\method\ (larger $\to$ higher het.) & \textbf{14.32}{\scriptsize$\pm$0.38} & \textbf{11.98}{\scriptsize$\pm$0.70} & \textbf{12.52}{\scriptsize$\pm$0.37} & \textbf{513.3}{\scriptsize$\pm$25.5} \\
Uniform & 13.82{\scriptsize$\pm$0.44} & 11.49{\scriptsize$\pm$0.56} & 11.99{\scriptsize$\pm$0.40} & 521.3{\scriptsize$\pm$28.8} \\
Inverse (larger $\to$ lower het.) & 12.67{\scriptsize$\pm$0.61} & 10.35{\scriptsize$\pm$1.71} & 10.89{\scriptsize$\pm$1.09} & 551.7{\scriptsize$\pm$13.5} \\
\bottomrule
\end{tabular*}
\end{table}

Table~\ref{tab:directionality} isolates the effect of allocation direction under the same fixed budget across 10 matched seeds. As in Table~\ref{tab:main_10seed}, these ablation statistics come from a separate dedicated 10-seed run, so small differences relative to Fig.~\ref{fig:main_bar} reflect run-to-run variability across independent training runs. Reversing the mapping so that clients with more heterogeneous data receive smaller subnets (Inverse) reduces mean accuracy from 14.32\% to 12.67\% and P10 from 12.52\% to 10.89\%. Paired one-sided $t$-tests confirm that \method\ significantly outperforms both Uniform ($\Delta{=}{+}0.50\%$, $t{=}4.90$, $p{<}0.001$) and Inverse ($\Delta{=}{+}1.65\%$, $t{=}8.72$, $p{<}0.001$) on mean accuracy.

Relative to Uniform, the Inverse mapping degrades mean accuracy by 1.15 percentage points and P10 by 1.10 percentage points. The Inverse policy also exhibits substantially higher variance on worst-client accuracy ($\pm$1.71\% versus $\pm$0.56\% for Uniform), indicating that assigning narrow subnets to heterogeneous clients makes training less stable. Mean perplexity increases from 513.3 (\method) to 551.7 (Inverse), corresponding to lower language-modeling quality. These results identify allocation direction as a major determinant of performance: allocating more width to more heterogeneous clients improves performance, while the reverse reduces it.

\subsection{Cross-Domain Image-Classification Results}

\begin{table}[t]
\centering
\caption{Cross-domain image-classification study on CIFAR-10 with 5 clients. Average and worst-client accuracies are computed at each client's assigned subnet width. Global accuracy at $r{=}0.5$ evaluates the global model at the common width ratio.}
\label{tab:cifar}
\scriptsize
\setlength{\tabcolsep}{2.5pt}
\renewcommand{\arraystretch}{1.3}
\begin{tabular*}{\columnwidth}{@{\extracolsep{\fill}}lccc@{}}
\toprule
Policy & Avg (\%) & Worst (\%) & Global at $r{=}0.5$ (\%) \\
\midrule
Uniform ($r{=}0.5$) & 68.05 & 59.00 & \textbf{70.12} \\
Size-only & 70.37 & 57.26 & 64.33 \\
HASA (visual complexity) & 71.71 & 59.83 & 62.06 \\
Inverse & \textbf{72.90} & \textbf{63.92} & 62.96 \\
Mixed (size + visual compl.) & 71.33 & 57.41 & 60.25 \\
\bottomrule
\end{tabular*}
\end{table}

The CIFAR-10 experiment is used as a cross-domain test of score alignment rather than as a second main benchmark. Table~\ref{tab:cifar} reports two complementary views of performance: client-level accuracy at each client's assigned subnet width, and global accuracy at the common width ratio $r{=}0.5$.

Under the common-width diagnostic, the uniform baseline achieves the best global test accuracy (70.12\%). In contrast, heterogeneity-based allocations improve average and, in some cases, worst-client accuracy at their assigned widths while performing worse on the common-width diagnostic. In this split, the Inverse policy---which assigns larger subnets to clients with lower visual complexity---is strongest on the client-level metrics, reaching 72.90\% average client accuracy and 63.92\% worst-client accuracy, while \method\ reaches 71.71\% and 59.83\%, respectively.

The CIFAR-10 result indicates that the score must reflect the form of heterogeneity that affects capacity demand. In the article-title benchmark, token-distribution JSD identifies domain-specific lexical shift and the higher-score-to-larger-width direction is beneficial. In the CIFAR-10 split, the visual-complexity score is misaligned with the dominant non-IID mechanism, which is closer to label-distribution skew, and the inverse direction performs better. Deployment should validate the proxy direction on a held-out validation signal when possible, or use a proxy that directly matches the expected non-IID structure, such as label-distribution divergence for label skew or frozen-encoder feature distance for feature skew.

\section{Discussion}

The two article-title protocols address different aspects of the evaluation. The original FedAvg comparison measures whether heterogeneity-aware allocation improves client-level performance under a fixed budget, while the selective-aggregation benchmark isolates the allocation rule against representative partial-training baselines. The absolute gains are modest but budget-neutral: \method\ changes which clients receive width, not the total nominal width. The mean-accuracy and P10 improvements are statistically supported; the worst-client gain is positive on average but not statistically significant at the 0.05 level.

The experiments represent controlled edge-node or cross-silo studies rather than large-scale cross-device deployments. Full participation isolates the allocation mechanism, whereas large-scale cross-device FL involves partial participation, availability bias, stragglers, and hard device-specific memory, latency, energy, and network constraints. In deployment, \method\ can be intersected with a feasibility cap $\bar{r}_i$ for each client, with the server enforcing the budget over the sampled cohort when participation is partial or availability-correlated.

The overhead metrics are analytical or simulated proxies. Weighted MACs estimate compute and active-submodel bytes estimate communication, but no device latency, memory, or energy measurements are reported. These proxies enable controlled budget matching; profiling representative edge devices and mapping measured constraints to client caps remain necessary for deployment-oriented evaluation.

The CIFAR-10 diagnostic shows that \method\ depends on proxy alignment. Token-distribution JSD is appropriate for the article-title benchmark because it measures lexical shift, whereas mean pixel standard deviation is poorly matched to label-distribution skew. These results support selecting proxies that match the expected non-IID mechanism, such as label-distribution divergence when labels are available or frozen-encoder feature distances for feature skew, and validating the allocation direction before deployment.

\section{Conclusion}

This paper proposes Heterogeneity-Aware Subnet Allocation (HASA), a train-only rule that assigns subnet widths based on client heterogeneity scores computed from local training data while enforcing a fixed size-weighted compute budget. This design enables budget-matched comparisons with alternative allocation policies. On an article-title next-word prediction benchmark with seven clients, HASA improves unweighted mean client test accuracy over uniform allocation across 10 matched seeds, increasing mean client test accuracy from 13.82 percent to 14.32 percent, and improves worst-client accuracy on average. In a matched-budget comparison with representative partial-training baselines, HASA achieves the strongest worst-client and tail-client accuracy on this benchmark. A directionality ablation shows that assigning smaller subnets to more heterogeneous clients degrades both mean and tail performance. A cross-domain image-classification study further shows that the effectiveness of heterogeneity-aware allocation depends on how well the heterogeneity score reflects clients' need for additional model width.

\section*{Acknowledgements}
\small
This work was supported in part by the Estonian Research Council grant PUT PRG1467 "CRASHLESS“, EU Grant Project 101160182 “TAICHIP“, by the Deutsche Forschungsgemeinschaft (DFG, German Research Foundation) – Project-ID "458578717", and by the Federal Ministry of Research, Technology and Space of Germany (BMFTR) for supporting Edge-Cloud AI for DIstributed Sensing and COmputing (AI-DISCO) project (Project-ID "16ME1127"), and UKRI-EPSRC grant No. EP/X035085/1.

\begingroup
\sloppy
\hbadness=10000
\bibliographystyle{IEEEtran}
\bibliography{refs}
\endgroup

\end{document}